\documentclass[conference]{IEEEtran}
\IEEEoverridecommandlockouts

\usepackage[switch]{lineno}

\usepackage[table]{xcolor}

\usepackage{csquotes}

\def\secref#1{Section~\ref{#1}}
\def\tabref#1{Table~\ref{#1}}
\def\figref#1{Figure~\ref{#1}}

\usepackage{stringstrings}
\def\pmp{$\pm$}
\newcommand\e[3]{\pmp #1.{#2}\textit{e}{-#3}}

\usepackage{multirow}
\usepackage{times}
\usepackage{soul}
\usepackage{url}
\usepackage[hidelinks]{hyperref}
\usepackage{booktabs}
\usepackage{algorithm}
\usepackage{algorithmic}
\usepackage{cite}
\usepackage{amsmath,amssymb,amsfonts}
\usepackage{algorithmic}
\usepackage{graphicx}
\usepackage{textcomp}
\usepackage{xcolor}
\def\BibTeX{{\rm B\kern-.05em{\sc i\kern-.025em b}\kern-.08em
  T\kern-.1667em\lower.7ex\hbox{E}\kern-.125emX}}
\begin{document}
\title{Don't Judge Me by My Face: An Indirect Adversarial Approach to Remove Sensitive Information From Multimodal Neural Representation in Asynchronous Job Video Interviews}

\author{\IEEEauthorblockN{Léo Hemamou\IEEEauthorrefmark{1}\IEEEauthorrefmark{2}\IEEEauthorrefmark{3}, Arthur Guillon\IEEEauthorrefmark{1}, Jean-Claude Martin \IEEEauthorrefmark{2} and Chloé Clavel\IEEEauthorrefmark{3}}
\IEEEauthorblockA{\IEEEauthorrefmark{1}EASYRECRUE, Paris, France \\
Email: \{l.hemamou,art.guillon\}@gmail.com},
\IEEEauthorblockA{\IEEEauthorrefmark{2}LIMSI-LISN, CNRS, Paris-Sud University, Paris-Saclay University / F-91405 Orsay, France\\
Email: Jean-Claude.Martin@limsi.fr}\IEEEauthorblockA{\IEEEauthorrefmark{3}Télécom-Paris, IP-Paris,/ F-75013 Paris, France\\
Email: chloe.clavel@telecom-paristech.fr}}


\maketitle

\begin{abstract}
Use of machine learning for automatic analysis of job interview videos has recently seen increased interest. Despite claims of fair output regarding sensitive information such as gender or ethnicity of the candidates, the current approaches rarely provide proof of unbiased decision-making, or that sensitive information is not used.
Recently, adversarial methods have been proved to effectively remove sensitive information from the latent representation of neural networks. However, these methods rely on the use of explicitly labeled protected variables (e.g. gender), which cannot be collected in the context of recruiting in some countries (e.g. France). 
In this article, we propose a new adversarial approach to remove sensitive information from the latent representation of neural networks without the need to collect any sensitive variable. Using only a few frames of the interview, we train our model to not be able to find the face of the candidate related to the job interview in the inner layers of the model.
This, in turn, allows us to remove relevant private information from these layers. Comparing our approach to a standard baseline on a public dataset with gender and ethnicity annotations, we show that it effectively removes sensitive information from the main network.
Moreover, to the best of our knowledge, this is the first application of adversarial techniques for obtaining a multimodal fair representation in the context of video job interviews. In summary, our contributions aim at improving fairness of the upcoming automatic systems processing videos of job interviews for equality in job selection.
\end{abstract}


\section{Introduction}

Job interviews are omnipresent in recruitment procedures. Although they are more generally conducted face-to-face, new forms of job interviews have emerged, such as the asynchronous video interview (AVI). For this form of interview, candidates connect to a web platform and answer a series of questions by recording themselves in a video monologue. Later on, recruiters have the opportunity to watch these videos, evaluate the candidates, and decide whether or not to invite them for a face-to-face interview. The growing interest in AVIs has naturally led to the development of algorithms and proprietary tools to automatically assess and rank candidates through the use of machine learning~\cite{Rasipuram2018AutomaticInteractions,Chen2017}. The adoption of this practice has met scepticism, both from lawmakers and the public, 
as sensitive information (e.g. gender or race) could be unintentionally used by these algorithms leading to unfair selection results. Vendors of algorithmic pre-employment assessment technologies generally claim unbiased predictions, but these claims are rarely backed with published studies or audits~\cite{Raghavan2019MitigatingPractices}. Moreover, those companies only guarantee fairness regarding the \textit{impact disparity} (i.e. if outcomes differ across subgroups) but not regarding the \textit{treatment disparity} (i.e. if subgroups are treated differently). It therefore seems necessary to improve these decision support models for a more accurate and unbiased process.

In this article, we aim to ensure equal treatment by preventing the automatic analysis system from using protected variables of the candidates. Despite these variables not being present in the input of the algorithm, we show that they leak into the hidden representation of the neural network. Sensitive information of the candidates can be retrieved from the inner layers of the network. Thus there is no way of guaranteeing that it is not used by the algorithm to predict the candidates’ interview performance or ‘hireability’ as it is commonly referrend to in the literature\cite{Nguyen2014HireBehavior,Rasipuram2018AutomaticInteractions,Hemamou2019HireNet:Interviews,Chen2017}. Recently, adversarial learning have been successfully proposed to obtain fair representations. To the initial prediction task is added a secondary adverserial task which consists of predicting the protected variables (e.g. gender and ethnicity of the candidates). The overall objective then becomes maximization of the hireability prediction task while minimizing the ability of the adversary branch to predict the protected variables. However, these techniques rely explicitly on these variables to train the adversary branch. This dependency raises two issues in the context of recruitment. Firstly, the collection of these variables is prohibited by the law in some countries. Secondly, recruitment suffers from multiple potential biases, like physical attractiveness, obesity or age, and explicit labeling of each protected variable by recruiters could become impossible in practice.

In \secref{sec:approach}, We propose, two indirect adversarial approaches, which do not explicitly rely on the protected variables of the candidate but only on images of their face, recorded during the video interview. These images are correlated with the protected information, but are available, as indeed automatic analysis of AVIs requires explicit consent from the candidates for the processing of their videos. By relying on a state of the art neural network designed to predict hireability, we propose two new architectures based on this approach.

In \secref{sec:experiments}, we compare the performance of our approach to standard adversarial techniques that explicitly use the protected variables. We show that it succeeds in removing this information from the latent layer of the network without degrading its performance on the main task or using sensitive candidate information. We believe that our contribution represents an important first step in the field of fairness for AVIs. While our data allows us to assess the impact of our approach only on gender and ethnicity features, we are of the opinion that the methods described are likely to deliver similar results for other types of biases.

\begin{figure*}[ht]
 \centering
 \includegraphics[width=0.60\textwidth]{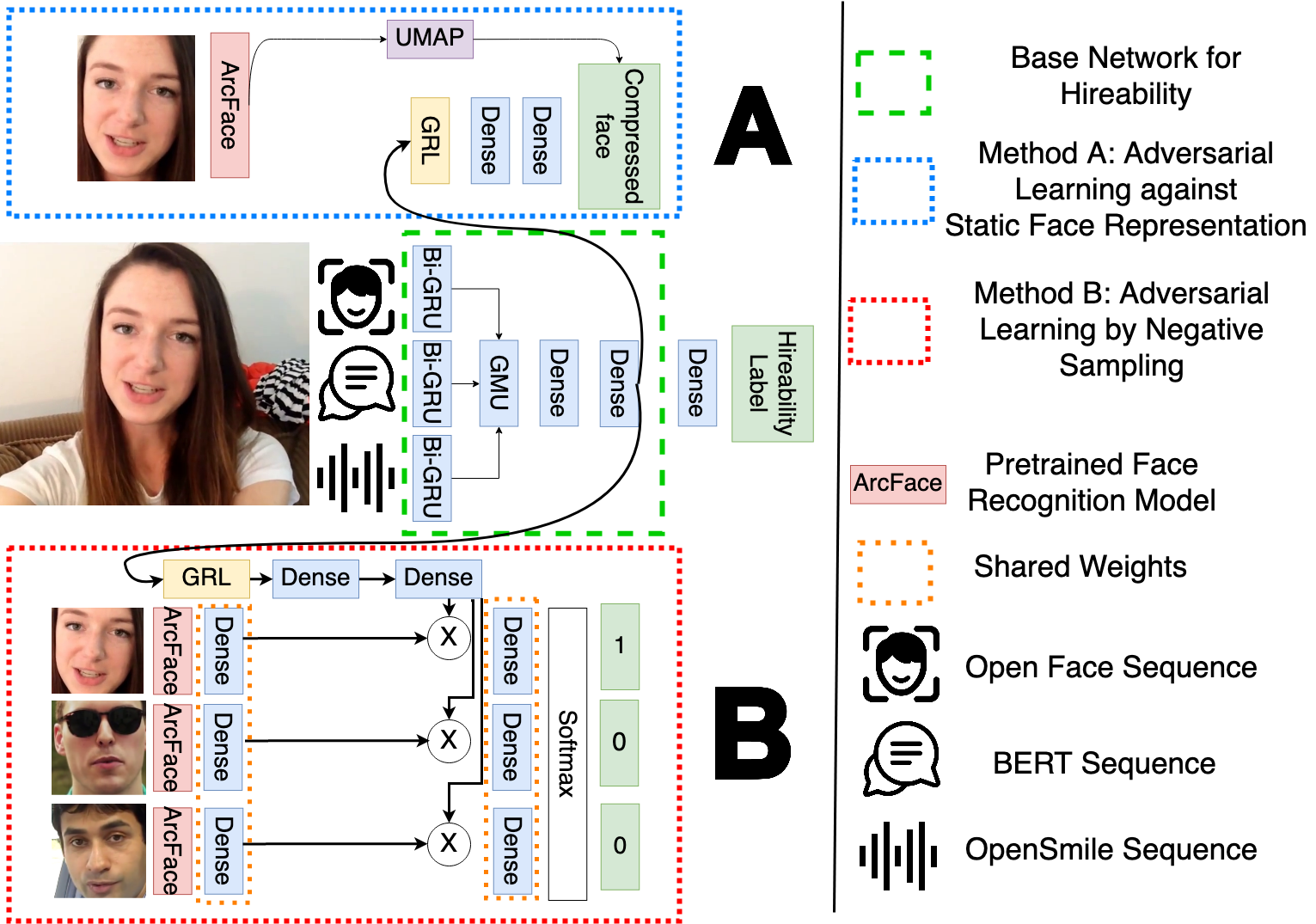}
 \caption{Architecture of the base network trained for hireability (green, middle of the figure) and the two proposed adversarial methods: static face representation (method A, blue, top) and negative sampling (B, red, bottom).}
  \label{fig:approach}
\end{figure*}

\section{Social context and related works}
\label{sec:related}


\subsection{Discrimination in Job Interviews}
\label{sec:fairness-recruiting}

Discrimination in job interviews is a recurring problem at the intersection of industrial and organizational psychology, social science and ethics.
Both the law and the theory of selection process indicate that decisions should only be made in relation to useful dimensions regarding the necessary skills of the job. However, literature has shown that factors such as gender, ethnicity, physical appearance or obesity do have an influence, even though taking a decision on these factors is expressly forbidden by law in both Europe and the United States \cite{2012TheSelection}. Moreover, in the context of AVIs, having access to the image of the candidates early in the selection process has a potential impact on increasing the influence of aesthetics~\cite{Torres2017AsynchronousSelection}, gender or ethnicity of the candidates~\cite{Kroll2016DiscriminationInterviews}. 
Approaches to mitigate discrimination in job interviews may differ from country to country \cite{Sanchez-Monedero2020WhatSystems}. For example,
in the US, the 4th/5th rule proposed by the EEOC (Equal Employment Opportunity Commission) states that the ratio of the least favored group compared to the most favored group must not be less than 0.8. In comparison, this rule cannot be used in some European countries where it is forbidden to collect sensitive attributes such as race \cite{Lieberman2001AStates}, preventing approaches based on statistical analysis.


These policies have an impact on the methods for measuring fairness used in the machine learning field~\cite{Corbett-Davies2018TheLearning}. For instance, \textit{classification parity} ensures that performance measures are equal between each of the protected groups, whereas \textit{anti-classification} ensures that no protected variable (race, gender, facial attractiveness, etc.) is used to make a decision.

\subsection{Automatic evaluation of AVIs}

Originally investigated in face-to-face settings~\cite{Naim2018AutomatedPerformance}, automatic analysis of video interviews has benefited from the emerging trend of AVIs. Indeed, the possibility of obtaining a large dataset as well as the structure of these interviews has allowed the use of machine learning~\cite{Rasipuram2018AutomaticInteractions,Chen2017} and more recently deep learning~\cite{Hemamou2019HireNet:Interviews,Escalante2020ModelingVideos,Leong2019ToAssessments,Singhania2020GradingConsiderations} to assess hireability. 
However, little work has been done to investigate the question of fairness in these systems~\cite{Raghavan2019MitigatingPractices}, which could produce biased outputs or unfair treatment towards minority groups. In fact, systematic bias could be the result of bias in the dataset, inadequate representation in the construction of the model, or under-representation of minority populations. Thus, some researchers have studied biases that might exist within the dataset~\cite{Escalante2020ModelingVideos,Leong2019ToAssessments} or in evaluating the fairness of the model's output~\cite{Singhania2020GradingConsiderations}. These studies are limited to ensuring that there is no bias in the system output or in the dataset, and they do not propose any method in the case of a biased pipeline. Furthermore, they do not ensure that sensitive information does not leak into the hiring decision of the system, which would contradict laws about equal treatment. 
In this paper, we adopt the framework of previous studies on the topic, including descriptor extraction through widely-used libraries (OpenFace~\cite{Baltrusaitis2018OpenFaceToolkit} and OpenSMILE~\cite{Eyben2013}) and obtaining a latent representation through a GRU-based neural architecture~\cite{cho2014learning}. We show that this methodology does indeed leaks sensitive information when inferring the hiring decision. 

\subsection{Fairness via adversarial methods}

Two major problems tackled by fairness in machine learning are group and individual fairness. \emph{Group fairness} designates the goal of producing equal or comparable results for populations belonging to different groups, according to some protected variable. Some approaches have been specifically designed to tackle this problem. For example, by taking into account the protected variables during inference \cite{Adel2019One-NetworkFairness}
, researchers obtain a perfect fair classifier in terms of group fairness. However, the application of such methods can be limited, as positive discrimination is forbidden in several countries (e.g. France, UK or Germany).

\emph{Individual fairness}, on the other hand, consists of giving a similar treatment to individuals that are similar regarding the task at hand. To this end, a recent family of approaches in machine learning focus on removing sensitive information from neural representation during training time through adversarial methods \cite{Xie2017ControllableLearning,Wang2019BalancedRepresentations,Adel2019One-NetworkFairness}. 
The idea is to obtain a representation without any information which could potentially lead to the recovering of the protected variables. Interestingly, these methodologies are also linked to privacy methods where network designers try to protect their system from attackers trying to retrieve personal information from latent representations. Thus, adversarial learning has been used in the context of speech processing \cite{Aloufi2020Privacy-preservingRepresentations}, deep visual recognition \cite{Wang2019BalancedRepresentations}, or multimodal (prosodic and verbal content) emotion recognition \cite{Jaiswal2020PrivacyRecognition}. However these approaches always rely on the explicit usage of the protected variables. In the setting of this paper, this would rely on asking the users for this information, a policy which would be inappropriate during a selection process or even illegal in several countries (collecting information about ethnicity is for example illegal in France \cite{Streiff-Fenart2012AFrance}).

In this article, we evaluate the utility of adversarial methods regarding protection of gender and ethnicity in the context of hireability prediction from multimodal monologue videos without the need of collecting additional sensitive information. 

\section{Proposed contribution: an indirect adversarial approach}
\label{sec:approach}

We present a new adversarial approach, available in two forms, to remove personal information of AVI candidates from the latent layer of a deep network trained for hireability. To this end, instead of using the candidates’ protected features directly, we use a representation of their face, which is extracted from the AVI. In this article, we integrate our approach with an architecture inspired by the state of the art of automatic AVI analysis. We first present the data used, the features and the base architecture in \secref{sec:materials}. The two new adversarial architectures are presented in \secref{sec:general-approach}. 

\subsection{Main task: base network for predicting hireability}
\label{sec:materials}


The data used in this article are videos from the dataset \enquote{ChaLearn First Impressions}
\cite{Escalante2020ModelingVideos}. To our knowledge, this dataset dataset is the only one which reproduces conditions of AVIs (monologue video), is annotated for hireability, provides sensitive information (gender and ethnicity) of the \enquote{candidates} and is publicly available. We also sought industrial partners for collaboration but found no company willing to explicitly collect sensitive information from real candidates.

\textbf{Dataset description.} The ChaLearn dataset consists of $n=10000$ video clips extracted from over
3000 YouTube videos of people facing the camera and speaking in English. The number of clips
appearing in the data for each video varies from 1 to 5 and their length is~15\,s. 
Each video was annotated for a binary hireability variable by Amazon Mechanical Turk workers, who were asked whether or not they would invite the persons to a job interview. Note that because annotators are not trained recruiters, we expect first impressions to have a greater impact on hireability inference, making it more difficult to remove sensitive information from the representation.
Although the ChaLearn dataset comes with a proposed 3-way split between training, validation and test sets, we found that 84\,\% of clips from the test set have at least one clip in the train set extracted from the same Youtube video. This overlap is potentially problematic, since it could allow classifiers to obtain good results on the hireability prediction task by learning speakers' specific features, e.g. facial or audio, a strategy that makes no sense from the point of view of recruitment. For this reason, we use our own 3-way split with no overlap. 

\textbf{Sensitive information.} Each video was annotated with gender (male or female) and ethnicity
(asian, caucasian or african-american)
. In this
article, we consider these annotations (gender and ethnicity) as the protected variables which
should not be used during the hireability inference. As such, they are only used for
the evaluation of the proposed algorithms. Table~\ref{tab:initial-bias} displays the average hireability label of each protected class in the complete dataset as well as in
all subsets, along with the distribution of the members of these classes. 

\begin{table*}
 \small
\centering
\begin{tabular}{|l|c|c|c|c|}
\hline
 Protected variable & Complete dataset & Training set (6991 clips) & Validation set (1448 clips) & Test set (1559 clips)\\
 \hline
 \hline
 Gender - Female & 0.560 &0.560 (3053) & 0.575 (702)& 0.541 (783)\\
 \hline
 Gender - Male & 0.495 &0.482 (3938) & 0.524 (746)& 0.519 (776)\\
 \hline
 Disparate Impact & 0.883 &0.860 & 0.911 & 0.959 \\
 \hline
 \hline
 Ethnicity - Asian &0.570 &0.584 (236)& 0.718 (32)& 0.444 (63)\\
 \hline
 Ethnicity - Caucasian & 0.541 &0.537 (5998)& 0.545 (1273)& 0.554 (1327)\\
 \hline
 Ethnicity - African-Americans & 0.434 &0.424 (757)& 0.559 (143)& 0.374 (171)\\
 \hline
 Disparate Impact &0.761 &0.726 & 0.759 & 0.675 \\
 \hline
 \end{tabular}%
 \normalsize
 \caption{Size and mean value of the hireability label for the ChaLearn First Impressions dataset, according to our training/validation/test sets and the classes formed by the two protected variables.}
 \label{tab:initial-bias}
\end{table*}

\textbf{Features for hireability prediction.} Following the methodology from automatic analysis of
interview literature~\cite{Hemamou2019HireNet:Interviews,Rasipuram2018AutomaticInteractions,Leong2019ToAssessments}, we use three different modalities: vocal cues of the video, facial
expressions of the candidate and verbal content. Each of these modalities is extracted from the
video as a time series. We extract audio frame-level features from each video using the OpenSmile ComParE~\cite{Schuller2013TheAutism} feature set, which is standard in the affective computing community. This results in a~$130$-d time series to represent audio. Frame-level facial expression features are extracted using the OpenFace
library~\cite{Baltrusaitis2018OpenFaceToolkit}. These include: position and rotation of the head,
presence and intensity of facial Action Units and direction of the gaze resulting in a vector of
dimension 52. Finally, as manual transcripts of each video clip are available, we compute a BERT~\cite{Devlin2019BERT:Understanding} encoding of each transcript using the second last layer of the base uncased BERT model provided by
HuggingFace. We build a representation of the transcript by sequencing the
embeddings word by word, obtaining a $768$-d time series.

\textbf{Base network for hireability.} Our approach was not designed with a specific architecture in mind and could be used on any deep learning algorithm trained for hireability. In conducting experiments for this article, we use an architecture adapted from that published previously under the name HireNet~\cite{Hemamou2019HireNet:Interviews}. Moreover, in addition to the monomodal networks from~\cite{Hemamou2019HireNet:Interviews}, we also propose to study the impact of our approach on a simple multimodal representation, by fusing the three modalities through the use of a Gated Multimodal Unit (GMU,~\cite{Arevalo2020GatedNetworks}). The GMU works by projecting all modalities on the same space and learning the contribution of each through a gated mechanism.

\figref{fig:approach} presents the base network as well as the two methods we propose. The base multimodal network is shown in the middle of the figure, which depicts the three modalities and the GMU, together with several dense layers and the binary hireability label. On the figure, the green box depicts the inner layers of the network. In the following, we denote by~$H$ the latent representation of the top layer of this subnetwork, and~$\theta_H$ the corresponding learned parameters. Likewise, we denote by~$\theta_D$ the parameters of the dense layer responsible for the hireability decision. Monomodal networks are not shown, but are obtained by removing the GMU layer and stacking the dense layers on top of the corresponding bidirectional GRU unit.

\textbf{Static representations of the candidates' faces.} Our approach uses static representation of
the candidates' face, captured during the AVI, to remove private information from the hidden layer
of the network. We use OpenFace confidence levels to
extract 5 frames where facial action units are minimal (to obtain the most neutral face).
Then, a face representation is obtained using a pre-trained neural network for face recognition
tasks, namely ArcFace \cite{Deng2019ArcFace:Recognition}. We obtain a 512-dimensional vector per
candidates by averaging face representations belonging to each candidate. In \figref{fig:approach}, usage of these static faces representations is shown by boxes labeled \enquote{ArcFace}.

\subsection{Indirect adversarial approach}
\label{sec:general-approach}

In the context of recruitment, the goal is to predict the hireability~$Y$ of the candidate from their video interview~$X$ without using protected features~$Z$. In the adversarial setting, this amounts to learning a latent representation~$H$ from which~$Z$ cannot be predicted. 
However, unlike in a classical adversarial framework,~$Z$ is not accessible. Instead, we have access to the candidate’s face W , which is correlated to~$Z$ but contains a lot of additional information. For this reason, it is necessary to project~$W$ on a simpler space that respects the protected classes formed by the variable~$Z$ (candidates of the same gender, ethnicity, glasses, hair color, tatoos, etc). In this paper, we propose two methods to compress~$W$ and learn an independent representation~$H$. 

Our two proposed methods employ adversarial training. More precisely, we aim to maximize performance for the hireability objective while minimizing an adversarial objective. The complete model is trained by optimizing the following min-max objective:
\begin{equation}
 \min_{\theta_H,\theta_D} \max_{\theta_A} \mathcal{L}_T(\theta_H,\theta_D)-\lambda \mathcal{L}_A(\theta_H,\theta_A)     
\end{equation}
where $\theta_H$ and~$\theta_D$ are the parameters of the base networks defined in the previous section, $\theta_A$ is the parameter of the adversarial branch, $\lambda \geq 0$ is a trade-off hyperparameter between the hireability objective and the adversarial objective, $\mathcal{L}_T$ is the loss function for the hireability prediction, i.e. binary cross-entropy, and~$\mathcal{L}_A$ is the loss function for the adversarial branch, detailed in the next sections.

Our adversarial approach is implemented through a Gradient Reversal Layer (GRL,~\cite{Ganin2015UnsupervisedBackpropagation}), which is a special layer with no weight vector. It behaves as the identity function during the forward pass, but negates the gradient during the backward pass, resulting in the network \emph{unlearning} the corresponding information. \figref{fig:approach} present both methods (top and bottom of the figure).

\textbf{First method: static representation of the faces.}
The first method we propose, shown at the top of \figref{fig:approach}, uses the previously computed ArcFace representation of the candidates' faces as a target for the adversarial branch. 
However, instead of using the raw ArcFace representation (a $512$-d vector), we extract a compressed representation $W'$ using the UMAP algorithm~\cite{McInnes2018UMAP:Projection}. UMAP is a dimensionality reduction method that projects the data while respecting local distances for similar points and global distances for dissimilar points. By preserving intra and inter-cluster distances, our goal is to obtain a low-level representation that respects the majority of inter-individual differences within the data, which we hope will coincide with the gender and ethnicity of the candidates. In this case, the adversarial objective function is the classical mean squared loss function :
\begin{equation}
\mathcal{L}_A = \frac{1}{N}\sum_{i=1}^N(W^{'}_i-f(H_i))^2  \label{eq}
\end{equation}

where $f$ denotes the dense layers on top of the GRL.


\textbf{Second method: negative sampling.}
We propose a second method, independent of the first, shown at the bottom of the \figref{fig:approach}. Rather than relying on a compressed representation of the candidates' faces, we sample~$k-1$ negative faces representation from other candidates. Like the previous method, a GRL is grafted just after the processing of the interview representation $H$. Then the task of the adversarial branches is to identify which face is associated with the interview representation $H$. More precisely we compute for each pair ${(H,W_l)}_{l=1..k}$ the value of interest $p_l$ :
\begin{equation}
    p_l= \frac{\exp( g(H,W_l))}{\sum_{j=1}^k \exp( g(H,W_j))}
\end{equation} 
where $g$ is a learnt function behaving as a similarity function between $H$ and $W_l$. Finally, we obtain the following adversarial loss function :
\begin{equation}
\mathcal{L}_A = \frac{1}{N}\sum_{i=1}^N\sum_{l=1}^k \delta_{il} \log(p_l) = \frac{1}{N} \sum_{i=1}^N \log(p_i)    
\end{equation} 

where~$\delta_{il} = 1$ if face~$l$ matches the \mbox{$i$th}~interview, $0$ otherwise.

In practice, as shown at the bottom of the \figref{fig:approach}, $g$ consists of three sub-components. The first one is the encoding of the faces (left of the frame on the figure). The second one is, like the previous method, a branch based on a GRL in charge of encoding the representation of the interviews. The third one consists of a Hadamard product between the face and the interview representation followed by a dense layer. We chose~$k$ equal to 5 resulting in batches of one positive and 4 negative pairs.

\section{Experiments}
\label{sec:experiments}

In order to assess our approach, we compare the algorithms from \secref{sec:approach} with a standard adversarial baseline which explicitly uses the protected variables of gender and ethnicity. 

\subsection{Evaluation metrics and baseline classifiers}
\label{sec:evaluation-metrics}

\textbf{Hireability metrics.} Evaluation metrics used for the hireability task are the accuracy (ACC) and the Area under the ROC curve (AUC). These evaluation metrics are well suited for binary classification and have been widely used in automatic analysis of AVIs~\cite{Chen2017,Escalante2020ModelingVideos}.

\textbf{Fairness metrics.} In order to assess the presence of sensitive information in the latent representation of the base network, we use a standard privacy-inspired metric: first, we train two \emph{diagnostic classifiers}, Logistic Regression (LR) and XGBoost~\cite{Chen2016XGBoost:System} (XGB) to recover the protected variable (gender or ethnicity) from the latent representation of the network, using the same training, validation and test sets. We use the AUC of these classifiers as a metric. In the case of a multi-class problem, we report the macro one-vs-rest AUC. This protocol is widely used in the privacy and fairness literature~\cite{Jaiswal2020PrivacyRecognition,Xie2017ControllableLearning}. Good performance of the diagnostic classifiers means that sensitive information is still contained in the latent representation.

The second metric we use is the Disparate Impact (DI):
\begin{equation}
DI = \frac{Pr(Y=1 | Z = unprivileged)}{Pr(Y=1 | Z = privileged)}
\end{equation}

Although it is not explicitly optimized by our proposed approach, it enables us to monitor any potential change in the DI as this metric is used in human resources and in the fair machine learning literature~\cite{Sanchez-Monedero2020WhatSystems}. A DI closer to 1 induces a fairer selection rate across groups.

\textbf{Standard supervised adversarial learning.} We compare our approach to a standard adversarial method which explicitly uses either gender or ethnicity variables. These baselines have the same architecture as the first method from \secref{sec:general-approach}, except that the adversarial head is replaced by a sigmoid (gender) or a softmax (ethnicity) to predict the class, and are called \enquote{Supervised} in the following.

\textbf{Size of the representations.} As pointed out in section \secref{sec:general-approach}, the dimensions of the compressed representation~$W$ could be critical. In order to investigate this aspect, we run our experiments by varying this dimension. Specifically, we evaluate both approaches by using either dimensions 2 (highly compressed representation) or 16 (low dimensional representation). 


\subsection{Experiments results}
\label{sec:expe-unsupervised}


\begin{table*}[ht]
\def\pmp{$\pm$}
\rowcolors{2}{gray!25}{white} 
\centering
{  \small
\begin{tabular}{lrrrrrr}
    \toprule  
    & \multicolumn{2}{c}{Hireability} & Gender & Ethnicity &
     DI Gender & DI Ethnicity \\
    \cmidrule{2-3}     
    \rowcolor{white}   
    & AUC & ACC & AUC & AUC  & & \\
    \midrule   
  Language modality\\   
  \hspace{1em} Unprotected & 0.613 & 0.584  & 0.584 & 0.500 & 0.984 & 0.867 \\
  \hspace{1em} Supervised gender &  0.611 \e{9}{8}{4} & 0.582 \e{6}{4}{4} & 0.567 \e{2}{5}{2} & - & 0.957 \e{1}{1}{2} & - \\
  \hspace{1em} Supervised ethnicity & 0.609 \e{1}{3}{5} & 0.576 \e{2}{9}{3} & - & 0.522 \e{1}{9}{2} & - & 0.832 \e{2}{7}{2} \\
  \hspace{1em} Static faces 2 & 0.611 \e{1}{0}{3} & 0.580 \e{9}{7}{4} & 0.578 \e{5}{0}{3} & 0.503 \e{6}{7}{3} & \textbf{0.986 \e{1}{2}{2}} & 0.818 \e{2}{9}{2} \\
  \hspace{1em} Static faces 16 & 0.612 \e{1}{1}{3} & 0.580 \e{2}{3}{3} & 0.573 \e{4}{0}{3} & 0.506 \e{1}{3}{2} & 0.963 \e{1}{0}{2} & 0.834 \e{1}{6}{2} 
  \\
  \hspace{1em} Negative sampling 2 & 0.604 \e{5}{8}{3} & 0.578 \e{1}{3}{3} & \textbf{0.527 \e{2}{2}{2}} & 0.527 \e{4}{9}{-3} & 0.974 \e{1}{0}{2} & 0.828 \e{1}{1}{1}\\ 
  \hspace{1em} Negative sampling 16 & 0.609 \e{1}{7}{3} & 0.578 \e{5}{8}{3} & 0.565 \e{2}{3}{2} & \textbf{0.500 \pmp 0.0}& 0.973 \e{2}{0}{2} & \textbf{0.845 \e{4}{1}{2}}
  \\
  Audio modality\\
  \hspace{1em} Unprotected & 0.695 & 0.642 & 0.850 & 0.506 & 0.837 & 0.913 \\
  \hspace{1em} Supervised gender & 0.692 \e{1}{3}{4} & 0.638 \e{6}{7}{3} & 0.584 \e{6}{7}{3} & - & 0.954 \e{8}{6}{3} & -\\
  \hspace{1em} Supervised ethnicity &  0.695 \e{4}{4}{4} & 0.640 \e{2}{3}{3} & - & 0.511 \e{7}{1}{3} & - & 0.892 \e{1}{1}{2} \\
  \hspace{1em}  Static faces 2 & 0.674 \e{5}{8}{3} & 0.597 \e{2}{5}{2} & \textbf{0.593 \e{1}{4}{2}} & \textbf{0.501 \e{2}{5}{2}} & 0.880 \e{3}{1}{2} & \textbf{0.946 \e{2}{0}{2}} \\
  \hspace{1em} Static faces 16 & 0.695 \e{5}{6}{4} & 0.642 \e{6}{4}{4} & 0.811 \e{2}{6}{2} & 0.505 \e{1}{2}{2} & 0.847 \e{3}{7}{3} & 0.903 \e{1}{9}{2} 
  \\
  \hspace{1em} Negative sampling 2 & 0.678 \e{4}{2}{3} & 0.645 \e{9}{6}{3} & \textbf{0.593 \e{4}{0}{2}} & 0.514 \e{8}{4}{3} & 0.947 \e{3}{7}{2} & 0.856 \e{3}{2}{2} \\
  \hspace{1em} Negative sampling 16 & 0.674 \e{1}{6}{2} & 0.633 \e{9}{8}{3} & 0.642 \e{6}{1}{2} & 0.519 \e{2}{2}{2} & \textbf{0.956 \e{5}{2}{2}} & 0.881 \e{4}{3}{2}
  \\
  Video modality\\
  \hspace{1em} Unprotected & 0.700 & 0.647 & 0.745 & 0.661 & 0.591 & 0.708 \\
  \hspace{1em} Supervised gender &  0.693 \e{3}{5}{3} & 0.647 \e{3}{2}{3} & 0.650 \e{1}{1}{2} & - & 0.769 \e{2}{0}{2} & -\\
  \hspace{1em} Supervised ethnicity & 0.698 \e{1}{2}{3} & 0.654 \e{3}{3}{3} & - & 0.520 \e{1}{7}{2} & - & 0.744 \e{3}{3}{3} \\
  \hspace{1em} Static faces 2 & 0.700 \e{5}{9}{4} & 0.650 \e{4}{0}{3} & 0.665 \e{5}{3}{3} & 0.646 \e{9}{1}{3} & 0.704 \e{1}{8}{2} & \textbf{0.864 \e{6}{7}{2}} \\    
  \hspace{1em} Static faces 16 & 0.699 \e{8}{3}{4} & 0.651 \e{1}{9}{3} & 0.708 \e{1}{6}{2} & 0.643 \e{1}{2}{3} & 0.636 \e{9}{2}{3} & 0.775 \e{7}{2}{3}\\ 
  \hspace{1em} Negative sampling 2 & 0.658 \e{6}{4}{3} & 0.613 \e{4}{3}{3} & 0.659 \e{1}{9}{2} & \textbf{0.585 \e{1}{3}{2}} & \textbf{0.971 \e{4}{5}{2}} & 0.745 \e{9}{0}{2} \\
  \hspace{1em} Negative sampling 16 & 0.693 \e{2}{2}{3} & 0.643 \e{1}{6}{3} & \textbf{0.656 \e{1}{3}{2}} & 0.630 \e{2}{2}{2} & 0.692 \e{2}{6}{2} & 0.707 \e{1}{5}{2}
  \\
  Multimodality\\
  \hspace{1em} Unprotected & 0.741 & 0.682 & 0.762 & 0.582 &  0.695 & 0.656 \\
  \hspace{1em} Supervised gender & 0.730 \e{5}{2}{3} & 0.675 \e{1}{1}{2} & 0.647 \e{5}{6}{2} & - & 0.851 \e{9}{5}{2} & -\\
  \hspace{1em} Supervised ethnicity & 0.731 \e{2}{7}{2} & 0.669 \e{5}{5}{3} & - & 0.538 \e{3}{2}{2} & - & 0.636 \e{3}{2}{2} \\
  \hspace{1em} Static faces 2 & 0.730 \e{1}{9}{3} & 0.667 \e{8}{4}{3} & 0.644 \e{6}{2}{2} & \textbf{0.510 \e{3}{0}{2}} & 0.847 \e{2}{9}{2} & 0.640 \e{7}{2}{2} \\
  \hspace{1em} Static faces 16 & 0.741 \e{1}{1}{3} & 0.683 \e{2}{0}{3} & 0.752 \e{2}{2}{2} & 0.552 \e{7}{2}{3} & 0.759 \e{3}{9}{3} & 0.652 \e{1}{1}{2}\\
  \hspace{1em} Negative sampling 2 & 0.706 \e{1}{8}{2} & 0.655 \e{1}{6}{2} & \textbf{0.615 \e{1}{7}{2}} & 0.535 \e{1}{4}{2} & \textbf{0.911 \e{6}{2}{2}} & \textbf{0.852 \e{2}{2}{2}}
  \\
  \hspace{1em} Negative sampling 16 & 0.718 \e{4}{2}{3} & 0.656 \e{9}{7}{4} & 0.649 \e{2}{3}{2} & \textbf{0.509 \e{1}{3}{2}} & 0.860 \e{1}{0}{2} & 0.772 \e{5}{7}{2} \\
  \bottomrule
\end{tabular}
}
\caption{Performance (AUC and ACC) of the deep networks for the hireability task, maximum performance (AUC) between the diagnostic classifiers (LR or XGB) for the identification of gender and ethnicity from the latent representation of the networks, and disparate impact (DI) for gender and ethnicity of the networks, for the four architectures.}
\label{tab:results}
\end{table*}

\tabref{tab:results} presents the results of the experiments, averaged over~$3$ runs. Proposed methods appear under the name \enquote{Static faces} and \enquote{Negative sampling}, followed by the size of the representation. The results are detailed for each modality and each hireability classifier, allowing comparison of the base \enquote{unprotected} network to the adversarially trained networks. The Hireability column gives the performance 
for the main task. Columns AUC Gender and AUC Ethnicity present the maximum performance of the diagnostic classifiers, LR or XGB, for the task of predicting gender and ethnicity from the latent representation of the networks. Columns DI Gender and DI Ethnicity give the disparate impact of each classifier for the base network and each adversarially trained classifier. For each modality and fairness metric, best scores for indirect methods are shown in bold. We discuss the results as a series of questions and answers:


\emph{Do automatic predictions amplify the bias present in human annotation?}
We compare the DI columns for \enquote{unprotected} networks to the DI displayed in \tabref{tab:initial-bias}, which is linked to the annotations of the dataset. 
Regarding gender, DI for audio, video and multimodal models is lower than with human annotations, hence the predictions are less fair than gold truth annotations. On the other hand, predictions based on language modality are fairer.
Regarding ethnicity, the language and audio models show a higher disparate impact than actual human annotations, whereas video and multimodal models seem to degrade disparate impact. Interestingly, the multimodal model is the worst model regarding disparate impact whereas it presents the best performance regarding hireability. 

\emph{Can we recognize gender and ethnicity given representations trained solely for hireability task?}
We discuss here the AUC scores of diagnostic classifiers for both gender and ethnicity on the unprotected networks. The high AUC values show that gender can be retrieved from the audio, video and multimodal models, but not from the language model. AUC scores for ethnicity are important only for video and, to a lesser extent, multimodal models, meaning that some information is retrievable. 
Finally, contrary to \cite{Jaiswal2020PrivacyRecognition}, our multimodal model does not show a higher leakage than our unimodal models.


\emph{What is the impact of supervised adversarial training on protected variables?}
As our work is the first to apply adversarial methods to AVIs, we comment on the effectiveness of classical adversarial methods. By comparing the AUC and accuracy values in the first columns between the unprotected method and the two supervised methods, we observe a marginal decrease in performance. For the audio, video and multimodal modalities, the decrease in AUC of the diagnostic classifiers shows that sensitive information is effectively removed from the latent representation. These scores provide a gold standard for the indirect methods we propose.

\emph{What is the impact of indirect adversarial methods on hireability performance and protected variables?}
We discuss here the columns Hireability, Gender and Ethnicity for the indirect methods we propose, Static faces and Negative sampling. For the audio modality, the indirect methods decrease the AUC of the diagnostic classifier for gender (except for the Static faces 16 method, discussed in the next question). This results in a score comparable to the supervised method in two cases, but also slightly affects the AUC for Hireability. Ethnicity is little affected, since the diagnostic classifiers have an AUC very close to 0.5 even for the Unprotected classifier. For the video modality, there is little gain for gender. 
However, for ethnicity, only the Negative sampling 2 method obtains an important drop in AUC from the diagnostic classifier, but this performance comes at the price of a drop in performance on the hireability prediction task. In the multimodal case, the indirect methods obtain good results with only a slight impact on the classifier's performance.



\emph{What is this impact of dimensionality of the compressed representation?}
In all columns, there are differences in outcome among the 4 indirect adversarial approaches. In particular, we notice that approaches using a 2-dimensional representation for faces generally perform better than the same 16-dimensional approach. The diagnostic AUCs of the Unprotected and the Static faces 16 methods are very close, showing that the latter is particularly inefficient. On the contrary, the Negative sampling 2 method obtains generally low AUC diagnostic scores, and performs better than the other approaches for gender in language, for ethnicity in video and for the multimodal modality in general. However, it is also the model which obtains the worst performance on the Hireability  task, particularly in the video and multimodal cases, reflecting the tradeoff between the two competing tasks.

\begin{figure}[ht]
 \centering
 \includegraphics[width=0.45\textwidth]{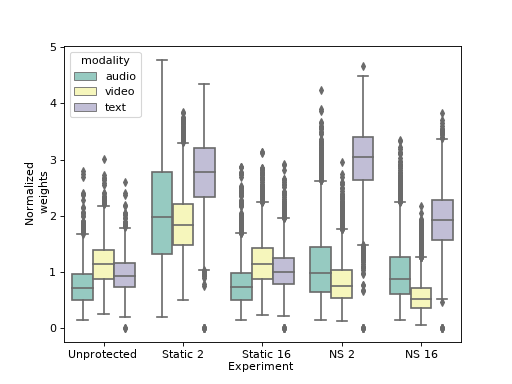}
 \caption{Boxplots of the contribution of each modality constituting multimodal vector through the GMU depending on the different adversarial training methods. \textit{NS} stands for Negative Sampling. The number stands for the dimension of the representation.}
  \label{fig:gmu-weights}
\end{figure}

\emph{What is the impact of indirect adversarial methods on the fusing step for the multimodal model?} 
Modalities are fused by the GMU through a weighted sum of each modality. As the model is simple enough we can investigate the norm of the vector for each modality, which reflects their degree of contribution to the multimodal representation. We display in \figref{fig:gmu-weights} the boxplots of the contribution of each modality to the multimodal vector depending on indirect adversarial methods used. Here, we observe that, for the unprotected model, the visual modality contributes the most, followed by the audio and language modalities. We then observe that the adversarial training does have an impact on the fusing step for the multimodal model especially for the Static Face 2, NS 2 and NS 16 methods. This influence is reflected in a greater contribution of the language modality and in a smaller contribution of the visual modality. Interestingly, the Static Face 16 method does not show this influence reinforcing our thinking that this method performs worse than the other three. Overall, we can assume that adversarial training tends to decrease the contribution of the most problematic modality and strengthens the contribution of the modality least likely to leak sensitive information.



\emph{Does our approach improve Disparate Impact as well?}
Although the approach we propose aims to reduce the potential disparity of treatment for automatic evaluation of AVIs, measuring the Disparate Impact shows that this improvement can also benefit group fairness. Thus, the last two columns of the \tabref{tab:results} show that adversarial methods often improve the Disparate Impact (with the exception of the Supervised Ethnicity method) compared to the Unprotected network, and that this increase is often even greater for indirect approaches that have succeeded in removing sensitive information from the latent representation. Both types of fairness can thus coexist, to a certain extent. 

\section{Conclusion and future work}
\label{sec:conclusion}



Guaranteeing equal treatment of candidates is crucial during a job interview. Here, we evaluate the potential for discrimination in a state-of-the-art hiring system on a public dataset. Despite not using sensitive features as input, we show that gender and ethnicity are retrievable from the latent representation of the system. 
Taking into account the fact that some countries prohibit the collection of this protected information in the context of recruiting, we use adversarial learning to remove sensitive information from the latent representation using accessible information which can be easily extracted from the video interviews as a proxy – the face of the candidates. Comparing this to more standard adversarial baselines, we show that our approach succeeds in removing protected information from the latent representations of the models. Moreover, we observe experimentally that the proposed algorithms generally improve the disparate impact even if group fairness is not targeted by our methods. 
Overall, we would like to reiterate that automatic hiring decisions should not be made without any human intervention and we believe that our work is a first step towards better addressing ethical issues in automatic analysis of AVI. In future work, we aim to test our approach on different datasets (e.g. more realistic or balanced), or other critical tasks such as automatic emotion recognition.


\section*{Acknowledgment}
I thank Peter Hagelund, Adrian Mihai and David Wright for their valuable assistance and William Coleman for proofreading.

\newpage

\bibliographystyle{IEEEtran}
\bibliography{references}

\newpage
\appendix

\section{Formal description of base networks, supervised adversarial method and indirect adversarial methods}

\subsection{Formalization}

In our model, we denote each instance by the triplet ${X,Y,Z}$ where $X$ denotes the candidate's answer, $Y$ the hireability label of this candidate and $Z$ the protected variables which must not be used for inference of $Y$.
Let $X_{\{L,V,A\}}$ $\in \mathbb{R}^{T_{\{L,V,A\}}  \times d_{\{L,V,A\}}} $ the sequence of low level descriptors describing the candidate's answer respectively from the three modalities: language~(L), video~(V) and audio~(A). $T_{m}$ and~$d_{m}$ represent the sequence length and feature dimension of the modality~$m$ $\in \{L,V,A\}$ as each modality has a different sampling frequency and feature dimension. We consider our problem for the main task as a classification problem: $Y$ is a binary variable (i.e. hirable or not hirable). Finally protected variable $Z$ is a categorical variable (e.g. gender or ethnicity).



\subsection{Architecture}
The goal of our model is to maximize the performance of the main task (e.g. hireability prediction) while minimizing the performance to predict $Z$. In that sense, our system is composed of three parts: a) the main network which encodes information from the multimodal answer $X$ to a latent representation $H$, b) the hireability classifier which classifies the answer based on the representation $H$ and c) the adversarial branch network which tries to predict $Z$ based on the representation $H$.

\textbf{Base  network  for  hireability.} Previous literature \cite{Hemamou2019HireNet:Interviews} suggests the relevance of recurrent neural networks for modeling the sequentiality of job video interviews. Inspired by such approaches we encode each modality
separately by a specific modality encoder in order to obtain a better representation of each element for the intra-modality representation~\cite{Poria2017Context-dependentVideos}. For that purpose, we use a reccurrent neural network component, namely a Bidirectional Gated Recurrent Unit (BiGRU). 
\begin{equation}
 z^m_{t} = BiGRU(x_t) , t \in [1,T_{m} ]
\end{equation}

In order to obtain a fixed size vector for each modality and to account for salient moments in the candidate's answer \cite{Hemamou2019HireNet:Interviews},  we use an additive attention mechanism, described by the following equations:
\begin{align}
 u^m_{t} &= \mathrm{tanh}(W^m_A z^m_{t} + b^m) \nonumber\\
  \alpha^m_t &= \frac{\mathrm{exp}(u^{m\top}_p {u^m_{t}} )}{\sum\nolimits_{t'} \mathrm{exp}(u^{m\top}_p {u^m_{t'}})} \nonumber\\
 o^m &= \sum\nolimits_{t} \alpha^m_t z^m_{t}
 \label{eq:low-level-output}
\end{align}
\noindent where~$W^m_A$ are weight matrices, $u^m_p$ and $b^m$ are weight vectors and $u^{m\top}_p$ denotes the transpose of  $u^m_p$ respectively for $m \in \{L,A,V\}$.

Previous work on job interview analysis using deep learning approaches \cite{Hemamou2019HireNet:Interviews,Leong2019ToAssessments} do not offer satisfactory multimodal representation of the inputs. We thus propose to improve the multimodal representation by fusing the three modalities through the use of a Gated Multimodal Unit (GMU,~\cite{Arevalo2020GatedNetworks}). The GMU works by projecting all modalities on the same space and learning the contribution of each through a gated mechanism. We propose to use it on top of the attention mechanism of each modality.
\begin{align*}
 \widetilde{o^a} &= \mathrm{tanh}(W_{Aproj} o^a) \nonumber\\
 \widetilde{o^l} &= \mathrm{tanh}(W_{Lproj} o^l) \nonumber\\
 \widetilde{o^v} &= \mathrm{tanh}(W_{Vproj} o^v) \nonumber\\
 \sigma^a &= \sigma(W_{Agating} [o^a,o^l,o^v]) \nonumber \\
 \sigma^v  &= \sigma(W_{Lgating} [o^a,o^l,o^v]) \nonumber \\
 \sigma^l  &= \sigma(W_{Vgating} [o^a,o^l,o^v])
 \end{align*}
 \begin{align}
 o^{multimodal}&=\sigma^a*\widetilde{o^a}+\sigma^l*\widetilde{o^l}+ \sigma^v*\widetilde{o^v}
 \label{eq:gmu}
\end{align}
\noindent where $W_{Aproj}$, $W_{Lproj}$, $W_{Vproj}$, $W_{Agating}$, $W_{Lgating}$, $W_{Vgating}$ are weight matrices, $\sigma$ is the sigmoid function and~$[x, y]$ is the concatenation of~$x$ and~$y$. Output of the multimodal unit is then given by~$o^{multimodal}$, which represents the multimodal information of the answer.

On top of the multimodal representation, we choose to use a simple combination of two dense layers, which is a simplification of the  architectures found in the literature and built on the structure of AVIs~\cite{Hemamou2019HireNet:Interviews,Leong2019ToAssessments}:
\begin{align}
  h_1 &= \mathrm{tanh}(W_1^\top o^{multimodal} + b_1) \label{eq:high-level-mono} \\
  H &= \mathrm{tanh}(W_2^\top h_1 + b_2)
  \label{eq:latent-representation}
\end{align}
\noindent where $W_1$ and $W_2$ are weight matrices and~$b_1$ and~$b_2$ are weight vectors. We denote by $\theta_H$ the parameters of the main network.

\textbf{Hireability classifier.} Once $H$ is obtained, we use it as representation in order to infer hireability:
\begin{equation}
 \widehat{Y} = \sigma(W_v H + b_v)
\end{equation}
where $W_v$ is a weight matrix and $b_v$ a weight vector.
We denote by $\theta_D$ parameters of the hireability classifier.

As the problem we are facing is that of a binary classification, the loss function of the main task is:
\[
 \mathcal{L}_T = -\frac{1}{N} \sum^N_{i=1} Y_i \log \widehat{Y_i} + (1-Y_i) \log (1-\widehat{Y_i})
\]
where $Y$ denotes the true labels of hireability of candidates.

\textbf{Supervised Adversarial Branch Network.}
\label{sec:adversarial-fairness} In order to force the network to forget sensitive information, we use an adversarial approach to train a second branch on top of the latent representation $H$, which learns to recover sensitive information about the candidates. This second branch is grafted on the main network through a Gradient Reversal Layer (GRL,~\cite{Ganin2015UnsupervisedBackpropagation}). A GRL is a special layer with no weight vector or matrix, which behaves as the identity function during the forward pass, but negates the gradient during the backward pass, resulting in the network \emph{unlearning} the corresponding information. In order to obtain a fairer representation, we decide to use a deeper network than the hireability branch. Also, as type of protected variables change, we use different activation or loss function. More precisely:
\[
  \widehat{Z} =
    \mathrm{softmax}\left(W_{4}\sigma(W_{3} H + b_{3})+ b_{4}\right) 
\]
where $W_{3}$, $W_{4}$ are trainable weight matrices and $b_{3}$,  $b_{4}$ are trainable vectors. We denote by~$\theta_A$ the parameters of the adversarial branch. Then, the loss of the adversarial branch is as follow :
\[
  \mathcal{L}_A =
    -\frac{1}{N}\sum_{i=1}^N\sum_{j=1}^C Z_{ij} \log(\widehat{Z}_{ij}) 
\]
where C denotes the number of classes for categorical $Z$ (e.g. 2 for gender and 3 for ethnicity in our case).

Following the notations used in the main article, the complete model is trained by optimizing the following min-max objective:
\[ \min_{\theta_H,\theta_D} \max_{\theta_A} \mathcal{L}_T(\theta_H,\theta_D)-\lambda \mathcal{L}_A(\theta_H,\theta_A) \]
where $\lambda \geq 0$  is a trade-off hyperparameter between the hireability objective and the adversarial objective.

\subsection{Indirect adversarial approach method 1 : static representation of the faces}

We describe in this subsection the function $f$ applied to the representation $H$ refered in section 3.2 "First method: static representation of the faces" of the main article.

$f$ consists in two dense layers, more precisely :

\[
f(H) = W_{6}(W_{5}H + b_{5})+ b_{6}
\]

where $W_{5}$, $W_{6}$ are trainable weight matrices and $b_{5}$,  $b_{6}$ are trainable vectors.

\subsection{Indirect adversarial approach method 1 : static representation of the faces}

We describe in this subsection the function $g$ applied to each pair ${(H,W_l)}_{l=1..k}$ referred in Section 3.2 "Second method: negative sampling" of the main article.

$g$ consists in three sub-components, more precisely :

\[
\widehat{W_l} = g_1(W_l) = tanh(W_{7}W_l + b_{7})
\]
$g_1$ is responsible for encoding the face represention in a low dimensional space.
$W_{7}$ is a trainable weight matrix and $b_{7}$ is a trainable vector.

\[
\widehat{H} = g_2(H) = tanh(W_{9}(W_{8}H + b_{8})+ b_{9})
\]
$g_2$ is responsible for encoding the interview represention.
$W_{8}$, $W_{9}$ are trainable weight matrices and $b_{8}$,  $b_{9}$ are trainable vectors.

\[
g(W_l,H) = g_3( \widehat{W_l},\widehat{H}) = W_{10}(\widehat{W_l} \odot \widehat{H} + b_{10})
\]
$g_3$ is responsible for estimate similarity measure between $\widehat{W_l}$ and $\widehat{H}$.
$W_{10}$ is a trainable weight matrix and $b_{10}$ is a trainable vector, $\odot$ stands for the Hadamard product (element-wise product).

\section{Training Strategy for Adversarial Learning}

We have followed this training strategy in order to train the model :

\begin{enumerate}

\item First we train the main network and the hirability branch ($\theta_H$ and $\theta_D$).
\item Then, we train only the adversarial branch ($\theta_A$)
\item The adversarial training is then organized by alternatively train the full network ($\theta_H$,$\theta_D$, and $\theta_A$) or only the adversarial branch ($\theta_A$)
    \begin{enumerate}
        \item Train the whole network (main network, hirability branch, adversarial branch on top of the GRL) for one epoch ($\theta_H$,$\theta_D$, and $\theta_A$)
        \item reset weights of the adversarial branch ($\theta_A$)
        \item train adversarial branch until any improvement on the validation loss  ($\theta_A$)
    \end{enumerate}
    We reiterate this loop until the validation loss defined by \[ \min_{\theta_H,\theta_D} \max_{\theta_A} \mathcal{L}_T(\theta_H,\theta_D)-\lambda \mathcal{L}_A(\theta_H,\theta_A) \] does not improve.

\end{enumerate}

\section{Hyper Parameters}

All the experiments, including the pretraining of base models for hireability, are ran~$3$ times and results are averaged. As mentionned in the attached code archive, all the experiments have been ran on a 32-core CPU server with 93 GB of RAM and P100 NVidia Tesla GPU with 16 GB of VRAM. The total runtime is between 3 weeks and 1 month.

\subsection{Hyperparameters of the model}

We describe here the hyperparameters (optimizer, layers sizes\dots) used for the architecture detailed in Section~4.2 in the main paper. In order to find the best hyperparameters, we conducted a grid search on the first task (hireability prediction), but observed little variation in the results. The final hyperparameters are displayed in Table~\ref{tab:hyperparameters}, using the same notations as in the paper.

Note that size of~$W_4$ and~$b_4$, which describe the final dense layer of the adversarial branch, vary depending on the protected variable: 1 for gender (sigmoid output), 3 for ethnicity (softmax output).

\begin{table}
\centering
\begin{tabular}{|p{3.5cm}|p{3.5cm}|}
\hline
Hyper parameter & Value or Dimensions \\
\hline
 Optimizer & Adam\\
 \hline

 Units of intra-modality Bi-GRU & 16\\
 \hline
  $W^m_A$, $u^m_p$ and $b^m$& [30,16],30,30\\
 \hline
  $W_{Aproj}$, $W_{Lproj}$, $W_{Vproj}$  & [48,16],[48,16],[48,16]\\
 \hline
  $W_{Agating}$, $W_{Lgating}$, $W_{Vgating}$ & [1,48],[1,48],[1,48]\\
 \hline
 $W_1$, $b_1$ & [16,48],16\\
 \hline
  $W_2$, $b_2$  &[16,16],16 \\
 \hline
  $W_v$, $b_v$ &[1,16],1 \\
 \hline
  $W_3$, $b_3$ & [30,16],30\\
 \hline
  $W_4$, $b_4$ depending on the output & [(1/3),30],(1/3)\\
  \hline
  $W_5$, $b_5$ & [30,16],30\\
  \hline
  $W_6$, $b_6$ depending on the output & [(2/16),30],(2/16)\\
  \hline
  $W_7$, $b_7$ depending on the hidden size chosen & [(2/16),512],(2/16)\\
  \hline
  $W_8$, $b_8$  & [32,16],32\\
  \hline
  $W_9$, $b_9$ depending on the output & [(2/16),32],(2/16)\\
  \hline
  $W_10$, $b_10$ depending on the hidden size chosen & [1,(2/16)],1\\
  \hline
  Batch size & 32\\
 \hline
  Regularization L2 & $10^{-4}$ \\
 \hline
  Dropout & $0.2$\\
 \hline
  Gradient Clip & 1.0\\
 \hline
 Learning rate joint training& 0.0001\\
 \hline
 Learning rate adversarial branch & 0.003\\
 \hline
 \end{tabular}%
 \caption{Hyper parameters used for training the model}
 \label{tab:hyperparameters}
 \end{table}

\begin{table*}
\centering
\begin{tabular}{|l|c|c|c|c|}
\hline
Protected variable & Language & Audio & Video & Multimodal \\
\hline
$\lambda$ Static faces 2  & 5.0  & 2.0 & 2.0 & 10.0 \\
 \hline
 $\lambda$ Static faces 16  & 10.0  & 5.0 & 2.0 & 1.0 \\
 \hline
 $\lambda$ Negative sampling 2  & 5.0  & 10.0 & 10.0 & 2.0 \\
 \hline
 $\lambda$ Negative sampling 16 & 10.0  & 5.0 & 2.0 & 1.0 \\
 \hline
$\lambda$ Supervised gender & 0.5 & 0.5 & 1.0 & 0.5\\
 \hline
$\lambda$ Supervised ethnicity & 10.0 & 0.5 & 1.0 & 10.0 \\
 \hline
 \end{tabular}%
 \caption{Summary of lambda chosen depending on modality and protected variables}
 \label{tab:lambdas}
 \end{table*}

During the adversarial training, we observed that the~$\lambda$ hyperparameter had a somewhat important effect on the performance. For the experiments, the grid used for the lambda value was~$\{0{.}5, 1, 2, 5, 10\}$. 
For a given modality, the selected~$\lambda$ parameter is
the one corresponding to the minimum final values of the loss function $\mathcal{L}_T-\mathcal{L}_A$ on the validation set, which
represents a trade-off between the hireability and the privacy-inducing tasks.   Table~\ref{tab:lambdas} displays the value chosen for all the models.

\subsection{Hyper parameters for diagnostic classifiers}

Two diagnostic classifiers are used to retrieve the protected variables from the network latent representation, Logistic Regression (LR) and XGBoost. During our initial experiments, we conducted a grid search on the hyperparameters of the two algorithms, but observed little variation on the performance of XGBoost classifier. Because the hyperparameter search was very extensive for this classifier, we decided to select a set of average values for the parameters: loss reduction split~$\gamma$ is fixed to~$0.1$, regularization parameter~$\alpha \in \{0.3, 0.5\}$ and number of estimators is~$500$. For LR, we observed more variation in the output depending on the regularization parameter. We thus run a hyperparameter search each time, selecting the best value in the range~$10^{-4}...10^4$, depending on the performance on the validation set, and both~$\ell_1$ and~$\ell_2$ norms were tested.

\section{Additional experiments on splits of the dataset}
\label{sec:splits}

\begin{table*}
 \small
\centering
\begin{tabular}{|l|c|c|c|c|}
\hline
 Protected variable & Complete dataset & Training set (6991 clips) & Validation set (1448 clips) & Test set (1559 clips)\\
 \hline
 \hline
 Gender - Female & 0.560 &0.560 (3053) &  0.575 (702)& 0.541 (783)\\
 \hline
 Gender - Male & 0.495 &0.482 (3938) &  0.524 (746)& 0.519 (776)\\
 \hline
 Disparate Impact & 0.883 &0.860 & 0.911 & 0.959 \\
 \hline
 \hline
 Ethnicity - Asian &0.570 &0.584 (236)&  0.718 (32)& 0.444 (63)\\
 \hline
 Ethnicity - Caucasian & 0.541 &0.537 (5998)&  0.545 (1273)& 0.554 (1327)\\
 \hline
 Ethnicity - Afro-Americans & 0.434 &0.424 (757)&  0.559 (143)& 0.374 (171)\\
 \hline
 Disparate Impact &0.761 &0.726 & 0.759 & 0.675 \\
 \hline
 \end{tabular}%
 \normalsize
 \caption{Summary of initial bias on the proposed split}
 \label{InitialBias}
\end{table*}

Although the ChaLearn dataset comes with a proposed 3-way split between training, validation and test sets, we found that 84\,\% of clips from the test set have at least one clip in the train set extracted from the same Youtube video. This overlap is potentially problematic, since it could allow classifiers to obtain good results on the hireability prediction task by learning speakers' specific features, e.g. facial or audio, a strategy that makes no sense from the point of view of recruitment. For this reason, we use our own 3-way split with no overlap. Table~\ref{InitialBias} presents the statistics of the protected variables for the new split.

\begin{table}[t!]
\centering
\begin{tabular}{|l|c|c||c|c|}
\hline
 & \multicolumn{2}{c||}{LR} & \multicolumn{2}{c|}{XGBoost}
 \\ \hline
 & ACC & AUC & ACC & AUC
 \\ \hline
Original split &  0.680 & 0.742 & 0.745 & 0.811\\ \hline
Proposed split &  0.650 & 0.695 & 0.622 & 0.672 \\ \hline
 \end{tabular}
\caption{Differences between the two splits of the dataset for hireability prediction using only static faces.}
\label{table:faces}
\end{table}



\begin{table*}[t!]
\centering
\begin{tabular}{|l|c|c||c|c||c|c||c|c|}
\hline
 & \multicolumn{2}{c||}{Language} & \multicolumn{2}{c||}{Audio} & \multicolumn{2}{c||}{Video}  & \multicolumn{2}{c|}{Multimodal}\\ \hline
 & ACC & AUC & ACC & AUC & ACC & AUC & ACC & AUC  \\ \hline
Original split &  0.601 & 0.607 & 0.664 & 0.721 & 0.686 & 0.751 & 0.701 & 0.766 \\ \hline
Proposed split &  0.584 & 0.610 & 0.642 & 0.695 & 0.647 & 0.700 & 0.682 & 0.741 \\ \hline
 \end{tabular}
\caption{Results for Hirability task}
\label{table:hire-auc}
\end{table*}


\subsection{Experiments on the original dataset}


As an illustration of the differences between the two sets, we conduct two experiments. 
Firstly,  we show that it is possible to obtain good results to the original ChaLearn challenge by using a naive baseline. For this, we use the original test set and, for each video, predict an hireability score by computing the average of the original hireability (binary) labels for candidates also present in training set. Missing values (candidates present in the test set only) are replaced with the average label on training and validation set. Comparing these values to the original labels of the test set, we obtain an AUC of~$0{.}797$ and an accuracy of~$0.7215$, which are comparable to the performance of most classifiers reported during the original challenge.
Secondly, we extract face representations of the candidates and train two classifiers (LR and XGBoost) on each of the splits. We report evaluation metrics regarding hireability in table~\ref{table:faces}. Classifiers trained on the original split have a higher score (AUC of 0.811) than the score obtained by classifiers trained and evaluated on the proposed split (AUC of 0.695): we interpret this result as proof that the sharing of videos occurring in the original split makes the task easier and could potentially reduce the hirability task to a person's identity recognition task.

\subsection{Hirability Performance depending on splits}

Finally, we report in Table~\ref{table:hire-auc} the performance of the 4 models studied in the paper on the original and proposed splits. We observe that the drop of performance between the two sets is not very important, except for the video modality. We interpret this as proof that the classifiers do not rely too much on information shared between the two sets and that they are pretty stable. This impression is strengthened by comparing these results to the Table~\ref{table:faces}. 

\newpage

\end{document}